\documentclass{article} 
\usepackage[T1]{fontenc}
\usepackage{tgtermes}
\usepackage{amsmath}

\usepackage{times}
\usepackage{amssymb}
\newcommand{\mathbold}[1]{\ensuremath{\boldsymbol{\mathbf{#1}}}}
\newcommand{\mbf}[1]{\ensuremath{\boldsymbol{\mathbf{#1}}}}

\usepackage[english]{babel}
\usepackage[parfill]{parskip}
\usepackage{afterpage}
\usepackage{framed}
\usepackage{xspace}

{\endMakeFramed}

\usepackage{lineno}

\usepackage{ragged2e}

\newcounter{parcount}

\DeclareRobustCommand{\parhead}[1]{\textbf{#1}~}
\DeclareRobustCommand{\PP}{\textcolor{Plum}{\P} }

\usepackage[usenames,dvipsnames]{xcolor}
\definecolor{shadecolor}{gray}{0.9}

\usepackage{graphicx}
\usepackage[labelfont=bf]{caption}
\usepackage[format=hang]{subcaption}

\usepackage{booktabs, array}

\usepackage[algoruled]{algorithm2e}
\usepackage{listings}
\usepackage{fancyvrb}
\fvset{fontsize=\small}

\usepackage{natbib}

\usepackage[colorlinks,linktoc=all]{hyperref}
\usepackage[all]{hypcap}
\hypersetup{citecolor=MidnightBlue}
\hypersetup{linkcolor=MidnightBlue}
\hypersetup{urlcolor=MidnightBlue}

\newcommand{\myeqp}[1]{\hyperref[eq:#1]{Eq.\ref*{eq:#1}}}
\newcommand{\mysec}[1]{\hyperref[sec:#1]{Section~\ref*{sec:#1}}}
\newcommand{\mytable}[1]{\hyperref[table:#1]{Table~\ref*{table:#1}}}
\newcommand{\myfig}[1]{\hyperref[fig:#1]{Figure~\ref*{fig:#1}}}
\newcommand{\myappendix}[1]{\hyperref[appendix:#1]{Appendix~\ref*{appendix:#1}}}
\newcommand{\myalg}[1]{\hyperref[alg:#1]{Algorithm~\ref*{alg:#1}}}
\newcommand{\mytheorem}[1]{\hyperref[theorem:#1]{Theorem~\ref*{theorem:#1}}}
\newcommand{\myfootnote}[1]{\hyperref[footnote:#1]{Footnote~\ref*{footnote:#1}}}

\usepackage
[acronym,smallcaps,nowarn,section,nonumberlist]{glossaries}
\glsdisablehyper{}

\lstdefinestyle{mystyle}{
    commentstyle=\color{OliveGreen},
    numberstyle=\tiny\color{black!60},
    stringstyle=\color{BrickRed},
    basicstyle=\ttfamily\scriptsize,
    breakatwhitespace=false,
    breaklines=true,
    captionpos=b,
    keepspaces=true,
    numbers=none,
    numbersep=5pt,
    showspaces=false,
    showstringspaces=false,
    showtabs=false,
    tabsize=2
}
\usepackage{iclr2016_conference}
\newacronym{ELBO}{elbo}{evidence lower bound}
\newacronym{KL}{kl}{Kullback-Leibler}

\newacronym{DEF}{def}{deep exponential family}
\newacronym{DLGM}{dlgm}{deep latent Gaussian model}
\newacronym{DRAW}{draw}{Deep Recurrent Attentive Writer}

\newacronym{MF}{mf}{mean-field}
\newacronym{VGP}{vgp}{variational Gaussian process}
\newacronym{BBVI}{bbvi}{black box variational inference}
\newacronym{ADVI}{advi}{automatic differentiation variational inference}
\newacronym{NF}{nf}{normalizing flows}
\newacronym{CVI}{cvi}{copula variational inference}
\newacronym{VAE}{vae}{variational autoencoder}
\newacronym{IWAE}{iwae}{importance weighted autoencoder}
\newacronym{NVIL}{nvil}{neural variational inference}
\newacronym{MIXTURE}{mixture}{}
\newacronym{DSVI}{dsvi}{}

\newacronym{VI}{vi}{variational inference}
\newacronym{EP}{ep}{expectation propagation}

\newacronym{GP}{gp}{Gaussian process}
\newacronym{GPLVM}{gplvm}{Gaussian process latent variable model}
\newacronym{ARD}{ard}{automatic relevance determination}
\newacronym{SVIGP}{svigp}{}
\newacronym{KISSGP}{kiss-gp}{}
\newacronym{FASTFOOD}{fastfood}{}

\glsunset{ADVI}
\glsunset{SVIGP}
\glsunset{MIXTURE}
\glsunset{CVI}
\glsunset{DSVI}
\glsunset{FASTFOOD}
\glsunset{KISSGP}
\glsunset{KL}

\newcommand{\mbx}{\mathbold{x}}

\newcommand{\mbz}{\mathbold{z}}

\newcommand{\mbI}{\mbf{I}}

\newcommand{\mbL}{\mbf{L}}

\newcommand{\mbtheta}{\mathbold{\theta}}
\newcommand{\mbepsilon}{\mathbold{\epsilon}}

\newcommand{\mbzero}{\mathbold{0}}
\newcommand{\mbxi}{\mathbold{\xi}}

\newcommand{\mbomega}{\mathbold{\omega}}

\newcommand{\mbsigma}{\mathbold{\sigma}}
\newcommand{\mbSigma}{\mathbold{\Sigma}}

\newcommand{\mblambda}{\mathbold{\lambda}}

\newcommand{\mbphi}{\mathbold{\phi}}

\newcommand{\mbmu}{\mathbold{\mu}}

\newcommand{\cL}{\mathcal{L}}
\newcommand{\cD}{\mathcal{D}}

\newcommand{\m}{\mathbf{m}}
\renewcommand{\S}{\mathbf{S}}

\newcommand{\f}{\mathbf{f}}

\newcommand{\x}{\mathbf{x}}

\newcommand{\mbs}{\mathbf{s}}
\newcommand{\mbt}{\mathbf{t}}
\newcommand{\Kxixi}{\mathbf{K}_{\mbxi\mbxi}}
\newcommand{\Kxis}{\mathbf{K}_{\mbxi s}}
\newcommand{\Kss}{\mathbf{K}_{ss}}

\renewcommand{\d}[1]{\ensuremath{\operatorname{d}\!{#1}}}
\newcommand{\g}{\,|\,}
\renewcommand{\gg}{\,\|\,}
\newcommand{\qvgp}{q_{\textsc{vgp}}}

\usepackage{tikz}
\usetikzlibrary{bayesnet}

\pgfdeclarelayer{edgelayer}
\pgfdeclarelayer{nodelayer}
\pgfsetlayers{edgelayer,nodelayer,main}

\definecolor{hexcolor0xbfbfbf}{rgb}{0.749,0.749,0.749}

\tikzset{>=latex}
\tikzstyle{none}   = [inner sep=0pt]

\tikzstyle{line}  = [ - ]
\tikzstyle{arrow}  = [ ->, shorten <=1pt, shorten >=1pt ]
\tikzstyle{ardash} = [ dotted, ->, shorten <=1pt, shorten >=1pt ]

\tikzstyle{empty}=[circle,opacity=0.0,text opacity=1.0,inner sep=0pt,minimum
width=0pt,minimum height=0pt]
\tikzstyle{box}=[rectangle,fill=White,draw=Black]
\tikzstyle{filled}=[circle,fill=hexcolor0xbfbfbf,draw=Black]
\tikzstyle{hollow}=[circle,fill=White,draw=Black]
\tikzstyle{param}=[rectangle,fill=Black,draw=Black,inner sep=0pt,minimum width=4pt,minimum height=4pt]

\definecolor{light}{RGB}{199, 153, 199}
\definecolor{dark}{RGB}{143, 39, 143}

\renewcommand\bibsection{%
   \section*{References}%
   \markboth{\MakeUppercase{\refname}}{\MakeUppercase{\refname}}%
  }%

\bibpunct{(}{)}{;}{a}{,}{,}
\usepackage{pifont}
\usepackage{amsthm}
\newtheorem{theorem}{Theorem}
\newtheorem{example}{Special Case}

\title{The Variational Gaussian Process}

\author{
Dustin Tran \\
Harvard University \\
\texttt{dtran@g.harvard.edu} \\
\AND
Rajesh Ranganath \\
Princeton University \\
\texttt{rajeshr@cs.princeton.edu}
\AND
David M. Blei \\
Columbia University \\
\texttt{david.blei@columbia.edu}
}

\iclrfinalcopy

\begin{document}

\maketitle

\begin{abstract}
Variational inference is a powerful tool for approximate inference,
and it
has been recently applied for representation learning with deep generative models.
We develop the \emph{\gls{VGP}}, a Bayesian nonparametric
variational family, which adapts its shape to match complex posterior distributions.
The \gls{VGP} generates approximate posterior samples by generating
latent inputs and warping them through random non-linear mappings; the
distribution over random mappings is
learned during inference, enabling the transformed outputs to
adapt to varying complexity. We prove a universal approximation theorem
for the \gls{VGP}, demonstrating its representative power for learning
any model. For
inference we present a variational objective inspired by
auto-encoders and perform black box inference over a wide class of
models.
The \gls{VGP} achieves new state-of-the-art results for
unsupervised learning, inferring models such as the
deep latent Gaussian model and the recently proposed DRAW.
\end{abstract}

%


\section{Introduction}

Variational inference is a powerful tool for approximate posterior
inference. The idea is to posit a family of distributions over the
latent variables and then find the member of that family closest to
the posterior.  Originally developed in the
1990s~\citep{hinton1993keeping,waterhouse1996bayesian,jordan1999introduction},
variational inference has enjoyed renewed interest around developing
scalable optimization for large datasets~\citep{hoffman2013stochastic},
deriving generic strategies for easily fitting many
models~\citep{ranganath2014black},
and applying neural networks as a flexible parametric family of
approximations~\citep{kingma2014autoencoding,rezende2014stochastic}.  This research has been particularly
successful for computing with deep Bayesian
models~\citep{neal1990learning,ranganath2015deep}, which
require inference of a complex posterior distribution~\citep{hinton2006fast}.

Classical variational inference typically uses the mean-field family,
where each latent variable is independent and governed by its own
variational distribution. While convenient, the
strong independence limits learning
deep representations of data. Newer research aims
toward richer families that allow dependencies among the latent variables.  One way to introduce dependence is to consider the
variational family itself as a model of the latent
variables~\citep{lawrence2000variational,ranganath2015hierarchical}. These
\textit{variational models} naturally extend to Bayesian hierarchies,
which retain the mean-field ``likelihood'' but introduce dependence
through variational latent variables.

In this paper we develop a powerful new variational
model---the~\glsreset{VGP}\gls{VGP}. The \gls{VGP} is a Bayesian
nonparametric variational model; its complexity grows efficiently and
towards \textit{any} distribution, adapting to the inference problem
at hand.
We highlight three main contributions of this work:
\begin{enumerate}
\item We prove a universal approximation theorem: under certain
  conditions, the \gls{VGP} can capture any continuous posterior
  distribution---it is a variational family that can be specified
  to be as expressive as needed.

\item We derive an efficient stochastic optimization algorithm for
  variational inference with the \gls{VGP}.  Our algorithm can be used in a wide class of
  models.  Inference with the \gls{VGP} is a black box variational
  method~\citep{ranganath2014black}.

\item We study the \gls{VGP} on standard benchmarks for unsupervised
  learning, applying it to perform inference in deep latent
  Gaussian models~\citep{rezende2014stochastic} and
  DRAW~\citep{gregor2015draw}, a latent attention model. For both models, we report
  the best results to date.
\end{enumerate}

\parhead{Technical summary.}
Generative models hypothesize a distribution
of observations $\mbx$ and latent variables $\mbz$, $p(\mbx,\mbz)$.
Variational inference posits a family of the latent variables $q(\mbz; \mblambda)$
and tries to find the variational parameters $\mblambda$ that are closest in
KL divergence to the posterior.  When we use a variational model,
$q(\mbz; \mblambda)$ itself might contain variational latent variables;
these are implicitly marginalized out in the variational family~\citep{ranganath2015hierarchical}.

The \gls{VGP} is a flexible variational
model. It draw inputs from a
simple distribution, warps those inputs through a non-linear mapping,
and then uses the output of the mapping to govern the distribution of
the latent variables $\mbz$. The non-linear mapping is itself a random
variable, constructed from a Gaussian process.  The
\gls{VGP} is inspired by ideas from both the Gaussian process latent variable
model~~\citep{lawrence2005probabilistic}
and Gaussian process
regression~\citep{rasmussen2006gaussian}.

The variational parameters of the \gls{VGP} are the kernel parameters
for the Gaussian process and a set of \emph{variational data}, which
are input-output pairs.  The variational data is crucial: it
anchors the non-linear mappings at given inputs and outputs. It is
through these parameters that the \gls{VGP} learns complex
representations. Finally, given data $\mbx$, we use stochastic optimization to
find the variational parameters that minimize the KL divergence to the model posterior.







\section{Variational Gaussian Process}

Variational models introduce latent variables to the variational
family, providing a rich construction for posterior
approximation~\citep{ranganath2015hierarchical}.  Here we introduce
the \acrfull{VGP}, a Bayesian nonparametric variational model that is
based on the \acrlong{GP}.  The \gls{GP} provides a class of latent
variables that lets us capture downstream distributions with varying
complexity.

We first review variational models and Gaussian processes.  We then
outline the mechanics of the \gls{VGP} and prove that it is a
universal approximator.

\subsection{Variational models}
\label{sec:background}

Let $p(\mbz\g\mbx)$ denote a posterior distribution over $d$ latent
variables $\mbz=(z_1,\ldots,z_d)$ conditioned on a data set
$\mbx$. For a family of distributions $q(\mbz; \mblambda)$
parameterized by $\mblambda$, variational inference seeks to minimize
the divergence
$\operatorname{KL}(q(\mbz; \mblambda)\gg p(\mbz\g\mbx))$.  This is
equivalent to maximizing the
\gls{ELBO}~\citep{wainwright2008graphical}.  The \gls{ELBO} can be
written as a sum of the expected log likelihood of the data and the KL
divergence between the variational distribution and the prior,
\begin{equation}
\label{eq:bound_autoencoder}
\cL = \mathbb{E}_{q(\mbz; \mblambda)}[\log p(\mbx\mid\mbz)] -
\operatorname{KL}(q(\mbz; \mblambda)\|p(\mbz)).
\end{equation}
Traditionally, variational inference considers a tractable family of
distributions with analytic forms for its density. A common
specification is a fully factorized distribution
$\prod_i q(z_i; \lambda_i)$, also known as the mean-field
family. While mean-field families lead to efficient computation,
they limit the expressiveness of the approximation.

The variational family of distributions can be interpreted as a model
of the latent variables $\mbz$, and it can be made richer by
introducing new latent variables. Hierarchical variational models
consider distributions specified by a variational prior of the
mean-field parameters $q(\mblambda;\mbtheta)$ and a factorized
``likelihood'' $\prod_i q(z_i\mid\lambda_i)$.  This specifies the
variational model,
\begin{equation}
\label{eq:hvm}
q(\mbz;\mbtheta)=\int
\Big[\prod_i q(z_i\mid\lambda_i)\Big]
q(\mblambda;\mbtheta)\d\mblambda,
\end{equation}
which is governed by prior hyperparameters $\mbtheta$.  Hierarchical
variational models are richer than classical variational
families---their expressiveness is determined by the complexity
of the prior $q(\mblambda)$. Many expressive variational
approximations can be viewed under this
construct~\citep{saul1996exploiting,jaakola1998improving,rezende2015variational,tran2015copula}.

\subsection{Gaussian Processes}
\label{sec:background:gaussian}

We now review the
\glsreset{GP}\gls{GP}~\citep{rasmussen2006gaussian}. Consider a data
set of $m$ source-target pairs $\cD=\{(\mbs_n,\mbt_n)\}_{n=1}^m$,
where each source $\mbs_n$ has $c$ covariates paired with a
multi-dimensional target $\mbt_n\in\mathbb{R}^{d}$. We aim to learn a
function over all source-target pairs,
$
\mbt_n = f(\mbs_n),
$
where $f:\mathbb{R}^c\to\mathbb{R}^d$ is unknown. Let the function
$f$ decouple as $f=(f_1,\ldots,f_d)$, where each
$f_i:\mathbb{R}^c\to\mathbb{R}$. \gls{GP} regression estimates the
functional form of $f$ by placing a prior,
\begin{equation*}
p(f) = \prod_{i=1}^d \mathcal{GP}(f_i; \mbzero, \Kss),
\end{equation*}
where $\Kss$ denotes a covariance function $k(\mbs,\mbs')$ evaluated over pairs
of inputs $\mbs,\mbs'\in\mathbb{R}^c$.
In this paper, we consider \gls{ARD} kernels
\begin{equation}
k(\mbs,\mbs') = \sigma^2_{\textsc{ard}} \exp\Big(
-\frac{1}{2}\sum_{j=1}^c \omega_j(s_j - s'_j)^2
\Big),
\label{eq:ard}
\end{equation}
with parameters
$\mbtheta=(\sigma^2_{\textsc{ard}},\omega_1,\ldots,\omega_c)$. The
weights $\omega_j$ tune the importance of each dimension.
They can be driven to zero during inference, leading to automatic
dimensionality reduction.

Given data $\cD$, the conditional distribution of the \gls{GP} forms a
distribution over mappings which interpolate between input-output
pairs,
\begin{equation}
  p(f \g \cD) = \prod_{i=1}^d \mathcal{GP}(f_i; \Kxis \Kss^{-1} \mbt_i, \Kxixi -
  \Kxis \Kss^{-1} \Kxis^\top).
  \label{eq:gp_posterior}
\end{equation}
Here, $\Kxis$ denotes the covariance function $k(\mbxi, \mbs)$ for an
input $\mbxi$ and over all data inputs $\mbs_n$, and $\mbt_i$
represents the $i^{th}$ output dimension.

\subsection{Variational Gaussian Processes}
\label{sec:vgp}

We describe the \acrfull{VGP}, a Bayesian nonparametric variational
model that admits arbitrary structures to match posterior
distributions. The \gls{VGP} generates $\mbz$ by generating latent
inputs, warping them with random non-linear mappings, and using the
warped inputs as parameters to a mean-field distribution.  The random
mappings are drawn conditional on ``variational data,'' which are
variational parameters. We will show that the
\gls{VGP} enables samples from the mean-field to follow arbitrarily
complex posteriors.

\begin{figure}[tb]
\begin{subfigure}[t]{0.5\columnwidth}
  \centering
  \begin{tikzpicture}[x=1.7cm,y=1.8cm]

  \node[obs]                  (z)      {$z_i$} ;
  \node[latent, left=of z]    (f)      {$f_i$} ;
  \node[latent, left=of f]    (xi)      {$\mbxi$} ;

  \factor[above=of f, xshift=-1.25cm] {theta} {$\mbtheta$} {} {};
  \factor[above=of f] {D} {\hspace{0.5cm}$\mathcal{D}=\{(\mbs,\mbt)\}$} {} {};

  \edge{xi}{f};
  \edge{f}{z};
  \edge{theta}{f};
  \edge{D}{f};

  \plate[inner sep=0.35cm, yshift=0.15cm,
    label={[xshift=-14pt,yshift=14pt]south east:$d$}] {plate1} {
    (z)(f)
  } {};

\end{tikzpicture}
  \vspace{1ex}
  \caption{\textsc{variational model}}
  \label{sub:vgp}
\end{subfigure}
\begin{subfigure}[t]{0.5\columnwidth}
  \centering
  \begin{tikzpicture}[x=1.7cm,y=1.8cm]

  \node[latent]               (z)      {$z_{i}$} ;
  \node[obs, right=of z]      (x)      {$\mbx$} ;

  \edge{z}{x};

  \plate[inner sep=0.35cm, yshift=0.15cm,
    label={[xshift=-14pt,yshift=14pt]south east:$d$}] {plate1} {
    (z)
  } {};

\end{tikzpicture}
  \vspace{1ex}
  \caption{\textsc{generative model}}
  \label{sub:black_box}
\end{subfigure}
\caption{\textbf{(a)} Graphical model of the \acrlong{VGP}.
The \gls{VGP} generates samples of latent variables $\mbz$ by
evaluating random non-linear mappings of latent inputs $\mbxi$, and
then drawing mean-field samples parameterized by the mapping.  These
latent variables aim to follow the posterior distribution for a
generative model \textbf{(b)}, conditioned on data $\mbx$.
}
\label{fig:vgp}
\end{figure}
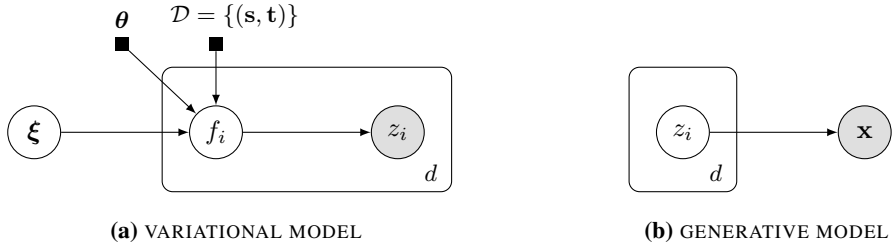

The \gls{VGP} specifies the following
generative process for posterior latent variables $\mbz$:
\begin{enumerate}
\item Draw latent input $\mbxi \in \mathbb{R}^c$:
$\mbxi\sim\mathcal{N}(\mbzero,\mbI).$
\item Draw non-linear mapping $f: \mathbb{R}^c\to\mathbb{R}^d$ conditioned on $\cD$:
$f\sim \prod_{i=1}^d\mathcal{GP}(\mbzero, \Kxixi) \g \cD.$
\item Draw approximate posterior samples $\mbz\in\operatorname{supp}(p)$:
$\mbz=(z_1,\ldots,z_d)\sim \prod_{i=1}^d
q(f_i(\mbxi) ).$
\end{enumerate}
\myfig{vgp} displays a graphical model for the \gls{VGP}.
Here, $\cD=\{(\mbs_n,\mbt_n)\}_{n=1}^m$ represents variational data, comprising
input-output pairs that are parameters to the variational
distribution.
Marginalizing over all latent inputs and non-linear mappings, the
\gls{VGP} is
\begin{equation}
\qvgp(\mbz; \mbtheta, \cD)
=
\iint
\left[
\prod_{i=1}^d q(z_i\g f_i(\mbxi))
\right]
\left[
\prod_{i=1}^d \mathcal{GP}(f_i; \mbzero, \Kxixi) \g \cD
\right]
\mathcal{N}(\mbxi; \mbzero,\mbI)
\d f \d\mbxi.
\label{eq:vgp}
\end{equation}
The \gls{VGP} is parameterized by kernel hyperparameters $\mbtheta$ and
variational data.

As a variational model, the \gls{VGP} forms an infinite ensemble of
mean-field distributions. A mean-field distribution is given in the
first term of the integrand above. It is
\emph{conditional} on a fixed function $f(\cdot)$ and input $\mbxi$;
the $d$ outputs $f_i(\mbxi)=\lambda_i$ are the mean-field's
parameters. The \gls{VGP} is a form of a hierarchical variational
model~(\myeqp{hvm})~\citep{ranganath2015hierarchical}. It places a continuous
Bayesian nonparametric prior over mean-field parameters.

Unlike the mean-field, the \gls{VGP} can capture correlation between the
latent variables. The reason is that it evaluates the $d$ independent
\gls{GP} draws at the same latent input $\mbxi$. This induces
correlation between their outputs, the mean-field parameters, and thus
also correlation between the latent variables. Further, the \gls{VGP} is
flexible. The complex non-linear mappings drawn from the \gls{GP}
allow it to capture complex discrete and continuous posteriors.

We emphasize that the \gls{VGP} needs variational data. Unlike typical
\gls{GP} regression, there are no observed data available to learn a
distribution over non-linear mappings of the latent variables $\mbz$.
Thus the "data" are variational parameters that appear in the
conditional distribution of $f$ in \myeqp{gp_posterior}. They anchor
the random non-linear mappings at certain input-ouput pairs. When
optimizing the \gls{VGP}, the learned variational data enables finds a
distribution of the latent variables that closely follows the
posterior.

\subsection{Universal approximation theorem}
\label{sec:vgp:universal}



To understand the capacity of the \gls{VGP} for representing complex
posterior distributions,
we analyze the role of the \acrlong{GP}. For simplicity,
suppose the latent variables $\mbz$ are real-valued, and the \gls{VGP}
treats the output of the function draws from the \gls{GP} as posterior
samples.
Consider the optimal
function $f^*$, which is the transformation such that when we draw
$\mbxi\sim\mathcal{N}(\mbzero,\mbI)$ and calculate $\mbz= f^*(\mbxi)$,
the resulting distribution of $\mbz$ \emph{is} the posterior distribution.

An explicit construction of $f^*$ exists if the dimension of the
latent input $\mbxi$ is equal to the number of latent variables. Let
$P^{-1}$ denote the inverse posterior CDF and $\Phi$ the standard
normal CDF. Using techniques common in copula
literature~\citep{nelsen2006introduction}, the optimal function is
\begin{equation*}
f^*(\mbxi) = P^{-1}(\Phi(\xi_1),\ldots,\Phi(\xi_d)).
\end{equation*}
Imagine generating samples $\mbz$ using this function. For latent
input $\mbxi\sim\mathcal{N}(\mbzero,\mbI)$, the standard normal CDF
$\Phi$ applies the probability integral transform: it squashes
$\xi_i$ such that its output $u_i=\Phi(\xi_i)$ is uniformly
distributed on $[0,1]$. The inverse posterior CDF then transforms the
uniform random variables $P^{-1}(u_1,\ldots,u_d)=\mbz$ to follow
the posterior. The function produces exact posterior samples.

In the \gls{VGP}, the random function interpolates the values in the variational data,
which are optimized to minimize the KL divergence.
Thus, during inference, the distribution of the \gls{GP} learns to concentrate around this
optimal function. This perspective provides intuition
behind the following result.


\newcommand{\limittheorem}{
Let $q(\mbz;\mbtheta, \cD)$ denote the \acrlong{VGP}. Consider a
posterior distribution $p(\mbz\g\mbx)$ with a finite number of latent
variables and continuous quantile function (inverse CDF). There exists
a sequence of parameters
$(\mbtheta_k, \cD_k)$ such that
\begin{equation*}
\lim_{k \to \infty} \operatorname{KL}(q(\mbz; \mbtheta_k, \cD_k) \gg p(\mbz\g\mbx))
= 0
.
\end{equation*}
}
\begin{theorem}
[Universal approximation]
\label{theorem:limit}
\limittheorem
\end{theorem}
See \myappendix{zero} for a proof. \mytheorem{limit} states that any
posterior distribution with strictly positive density
can be represented by a \gls{VGP}.
Thus the \gls{VGP} is a
flexible model for learning posterior distributions.

\section{Black box inference}
\label{sec:bbi}

We derive an algorithm for black box inference over a wide
class of generative models.

\subsection{Variational objective}
\label{sec:bbi:model}

\begin{figure}[tb]
  \centering
  \begin{tikzpicture}
  \pgfmathsetseed{3}
  \node at (0,0) {
    \begin{tikzpicture}[thick]
    \draw plot [smooth cycle, samples=8,domain={1:8}] (\x*360/8+5*rnd:0.5cm+1.5cm*rnd) node at (0,0) {$\mathcal{R}$};
    \end{tikzpicture}
  };
  \path[->,thick] (1.75,0) edge [bend left] node[above,align=center,yshift=0.125cm] {auxiliary\\inference} (3.25,0);
  \pgfmathsetseed{11}
  \node at (5,0) {
    \begin{tikzpicture}[thick]
    \draw plot [smooth cycle, samples=8,domain={1:8}] (\x*360/8+5*rnd:0.5cm+1.5cm*rnd) node at (0,0) {$\mathcal{Q}$};
    \end{tikzpicture}
  };
  \path[->,thick] (6.75,0) edge [bend left] node[above,align=center,yshift=0.125cm] {variational\\inference} (8.25,0);
  \pgfmathsetseed{9}
  \node at (10,0) {
    \begin{tikzpicture}[thick]
    \draw plot [smooth cycle, samples=8,domain={1:8}] (\x*360/8+5*rnd:0.5cm+1.5cm*rnd) node at (0,0) {$\mathcal{P}$};
    \end{tikzpicture}
  };
\end{tikzpicture}
\caption{Sequence of domain mappings during inference, from
variational latent variable space $\mathcal{R}$ to posterior latent
variable space $\mathcal{Q}$ to data space $\mathcal{P}$. We perform
variational inference in the posterior space and auxiliary
inference in the variational space.}
\label{fig:mappings}
\end{figure}
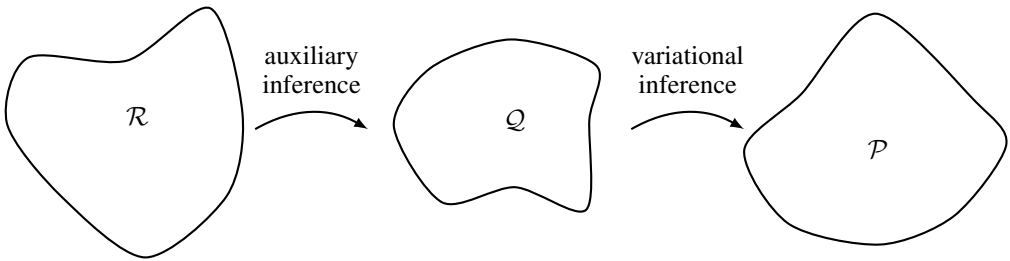

The original \gls{ELBO} (\myeqp{bound_autoencoder}) is
analytically intractable due to the log density,
$\log \qvgp(\mbz)$ (\myeqp{vgp}). To address this, we present a tractable variational
objective inspired by auto-encoders~\citep{kingma2014autoencoding}.

A tractable lower bound to the model
evidence $\log p(\mbx)$ can be derived by subtracting an expected KL
divergence term from the \glsunset{ELBO}\gls{ELBO},
\begin{equation*}
\log p(\mbx)
\ge
\mathbb{E}_{\qvgp}[\log p(\mbx\g\mbz)]
- \operatorname{KL}(\qvgp(\mbz)\|p(\mbz))
- \mathbb{E}_{\qvgp}\Big[
\operatorname{KL}(q(\mbxi,f\g\mbz)\|r(\mbxi,f\g\mbz))
\Big],
\end{equation*}
where $r(\mbxi,f\g\mbz)$ is an auxiliary model (we describe $r$ in
the next subsection).
Various versions of this objective have been considered in the
literature~\citep{jaakola1998improving,agakov2004auxiliary}, and it
has been recently revisited by \citet{salimans2015markov} and
\citet{ranganath2015hierarchical}.
We perform variational inference
in the posterior latent variable space, minimizing
$\operatorname{KL}(q\|p)$ to learn the variational
model; for this to occur we perform auxiliary inference in the
variational latent variable space, minimizing
$\operatorname{KL}(q\|r)$ to learn an auxiliary model.
See \myfig{mappings}.

Unlike previous approaches, we rewrite this variational objective to
connect to auto-encoders:
\begin{align}
\label{eq:variational_objective}
\begin{split}
\widetilde{\cL}(\mbtheta,\mbphi)
&
=
\mathbb{E}_{\qvgp}[\log p(\mbx\mid\mbz)]
-
\mathbb{E}_{\qvgp}\Big[
\operatorname{KL}(q(\mbz\g f(\mbxi))\|p(\mbz))
\Big]
\\
&
\quad
-
\mathbb{E}_{\qvgp}\Big[
\operatorname{KL}(q(f\g\mbxi;\mbtheta)\|r(f\g\mbxi,\mbz;\mbphi))
+
\log q(\mbxi) - \log r(\mbxi\g\mbz)
\Big],
\end{split}
\end{align}
where the KL divergences are now taken over tractable distributions
(see \myappendix{variational}).
In auto-encoder parlance, we maximize the expected negative
reconstruction error, regularized by two terms: an expected divergence between
the variational model and the original model's prior, and an expected
divergence between the auxiliary model and the variational model's
prior. This is simply a nested instantiation of the variational
auto-encoder bound~\citep{kingma2014autoencoding}: a
divergence between the inference model and a prior is taken as
regularizers on both the posterior and variational spaces. This
interpretation justifies the previously proposed
bound for variational models;
as we shall see, it also enables lower variance gradients during
stochastic optimization.

\subsection{Auto-encoding variational models}

An inference network provide a flexible parameterization of
approximating distributions as used in Helmholtz
machines~\citep{hinton1994autoencoders}, deep Boltzmann machines~\citep{salakhutdinov2010efficient}, and variational
auto-encoders~\citep{kingma2014autoencoding,rezende2014stochastic}.
It replaces local variational parameters with global parameters
coming from a neural network. For latent variables $\mbz_n$
(which correspond to a data point $\mbx_n$), an inference network
specifies a neural network which takes $\mbx_n$ as input and its local
variational parameters $\mblambda_n$ as output.  This amortizes
inference by only defining a set of global parameters.


To auto-encode the \gls{VGP} we specify inference networks to
parameterize both the variational and auxiliary models:
\begin{equation*}
\mbx_n\mapsto q(\mbz_n\g\mbx_n; \mbtheta_n),
\qquad
\mbx_n,\mbz_n\mapsto r(\mbxi_n,f_n\g\mbx_n,\mbz_n;
\mbphi_n).
\end{equation*}
Formally, the output of these mappings are the parameters $\mbtheta_n$ and
$\mbphi_n$ respectively. We write the output as distributions above to emphasize
that these mappings are a (global) parameterization of the variational
model $q$ and auxiliary model $r$.
The local variational parameters $\mbtheta_n$ for $q$ are the
variational data $\cD_n$. The auxiliary model $r$ is specified as a fully factorized
Gaussian with local variational parameters
$\mbphi_n=(\mbmu_n\in\mathbb{R}^{c+d}$,
$\mbsigma^2_n\in\mathbb{R}^{c+d})$.  \footnote{We let the kernel
hyperparameters of the \gls{VGP} be fixed across data points.
Note also that unique from other auto-encoder approaches, we let $r$'s
inference network take both $\mbx_n$ and $\mbz_n$ as input: this
avoids an explicit specification of the conditional distribution
$r(\mbepsilon,f\g\mbz)$, which may be difficult to model.
This idea was first suggested (but not implemented) in
\citet{ranganath2015hierarchical}.
}

\subsection{Stochastic optimization}
\label{sec:bbi:gradient}

We maximize the variational objective
$\widetilde{\cL}(\mbtheta,\mbphi)$ over both $\mbtheta$ and $\mbphi$, where
$\mbtheta$ newly denotes both the kernel hyperparameters and
the inference network's parameters for the \gls{VGP}, and $\mbphi$
denotes the
inference network's parameters for the auxiliary model.
%
Following the black box methods, we write
the gradient as an expectation and apply stochastic
approximations~\citep{robbins1951stochastic}, sampling from the
variational model and evaluating noisy gradients.

First, we reduce variance of the stochastic gradients by analytically
deriving any tractable expectations.
The KL divergence between
$q(\mbz\g f(\mbxi))$ and $p(\mbz)$ is commonly used to reduce
variance in traditional variational auto-encoders: it is analytic for
deep generative models such as the deep latent Gaussian
model~\citep{rezende2014stochastic} and deep recurrent
attentive writer~\citep{gregor2015draw}.
The KL divergence between
$r(f\g\mbxi,\mbz)$ and $q(f\g\mbxi)$ is analytic as the distributions
are both Gaussian. The difference $\log q(\mbxi) - \log
r(\mbxi\g\mbz)$ is simply a difference of Gaussian log densities.
See \myappendix{variational} for more details.

To derive black box gradients, we can first reparameterize the
\gls{VGP}, separating noise generation of samples from
the parameters in its generative
process~\citep{kingma2014autoencoding,rezende2014stochastic}. The
\gls{GP} easily enables reparameterization:
%
%
for latent inputs $\mbxi\sim\mathcal{N}(\mbzero,\mbI)$,
the transformation
$\f(\mbxi;\mbtheta)=\mbL\mbxi + \Kxis \Kss^{-1} \mbt_i$
is a location-scale transform,
where $\mbL\mbL^\top = \Kxixi - \Kxis \Kss^{-1} \Kxis^\top$.
This is equivalent to evaluating $\mbxi$ with a random mapping
from the \gls{GP}.
Suppose the mean-field $q(\mbz\g f(\mbxi))$ is also reparameterizable,
and let $\mbepsilon\sim w$ such that $\mbz(\mbepsilon;\f)$ is a
function of $\mbxi$ whose output $\mbz\sim q(\mbz\g f(\mbxi))$.
This two-level reparameterization is equivalent to the generative
process for $\mbz$ outlined in \mysec{vgp}.


We now rewrite the variational objective as
\begin{align}
\label{eq:reparam_objective}
\widetilde{\cL}(\mbtheta,\mbphi)
&
=
\mathbb{E}_{\mathcal{N}(\mbxi)}\Big[\mathbb{E}_{w(\mbepsilon)}\Big[
\log p(\mbx\mid\mbz(\mbepsilon;\f))
\Big]
-
\operatorname{KL}(q(\mbz\g \f)\|p(\mbz))
\Big]
\\
&
\quad
-
\mathbb{E}_{\mathcal{N}(\mbxi)}\Big[\mathbb{E}_{w(\mbepsilon)}\Big[
\operatorname{KL}(q(f\g\mbxi;\mbtheta)\|r(f\g\mbxi,\mbz(\mbepsilon;\f);\mbphi))
+
\log q(\mbxi) - \log r(\mbxi\g\mbz(\mbepsilon;\f))
\Big]\Big].
\nonumber
\end{align}
\myeqp{reparam_objective} enables gradients to move inside the expectations and backpropagate over
the nested reparameterization. Thus we can take unbiased stochastic
gradients, which exhibit low variance due to both the analytic
KL terms and reparameterization.
The gradients are derived in
\myappendix{gradients}, including the case when the first KL is
analytically intractable.

\begin{algorithm}[t]
  \caption{Black box inference with a \acrlong{VGP}}
  \SetAlgoLined
  \DontPrintSemicolon
  \BlankLine
  \KwIn{Model $p(\mbx,\mbz)$, Mean-field family $\prod_i q(\mbz_i\g
  f_i(\mbxi))$.}
  \BlankLine
  \textbf{Output}: Variational and auxiliary parameters
  $(\mbtheta,\mbphi)$.\;
  \BlankLine
  Initialize $(\mbtheta,\mbphi)$ randomly.
  \BlankLine
  \While{\textnormal{not converged}}{
    \BlankLine
    Draw noise samples $\mbxi\sim\mathcal{N}(\mbzero,\mbI)$,
    $\mbepsilon\sim w$.\;
    \BlankLine
    Parameterize variational samples
    $\mbz = \mbz(\mbepsilon;f(\mbxi))$, $f(\mbxi)=\f(\mbxi;\mbtheta)$.
    \BlankLine
    Update $(\mbtheta,\mbphi)$ with stochastic gradients
    $\nabla_{\mbtheta}\widetilde{\cL}$,
    $\nabla_{\mbphi}\widetilde{\cL}$.
  }
  \label{alg:vgp}
\end{algorithm}

We outline the method in \myalg{vgp}. For massive data, we
apply subsampling on
$\mbx$~\citep{hoffman2013stochastic}. For gradients of the model
log-likelihood, we employ convenient differentiation tools such as
those in Stan and Theano~\citep{carpenter2015stan,bergstra2010theano}.
For non-differentiable latent variables $\mbz$, or mean-field
distributions without efficient reparameterizations,
we apply the black box gradient estimator from \citet{ranganath2014black} to take gradients of the inner expectation.

\subsection{Computational and storage complexity}

The algorithm
has $\mathcal{O}(d+m^3+LH^2)$ complexity, where $d$ is the number of
latent variables, $m$ is the size of the variational data, and $L$
is the number of layers of the neural networks with $H$ the average
hidden layer size.
In particular, the algorithm is linear in the number of latent
variables, which is competitive with other variational inference methods.
%
%
%
%
The number of variational and auxiliary parameters has
$\mathcal{O}(c+LH)$ complexity; this complexity comes from storing the kernel
hyperparameters and the neural network parameters.
%

Unlike most \gls{GP} literature, we require no low rank constraints,
such as the use of inducing variables
for scalable computation~\citep{quinonero2005unifying}. The variational data serve a similar
purpose,
but inducing variables reduce the rank of a (fixed) kernel matrix; the
variational data directly determine the kernel matrix and thus the
kernel matrix is not fixed.
Although we haven't found it necessary in practice, see
\myappendix{size} for scaling the size of variational data.

%
%
%


\section{Related work}
\label{sec:related}

Recently, there has been interest in applying parametric
transformations for approximate inference. Parametric transformations of
random variables induce a density in the transformed space, with a
Jacobian determinant that accounts for how the transformation warps
unit volumes. \citet{kucukelbir2016automatic} consider this viewpoint
for automating inference, in which they posit a transformation from
the standard normal to a possibly constrained latent variable space.
In general, however, calculating the Jacobian determinant incurs a
costly $\mathcal{O}(d^3)$ complexity, cubic in the number of latent
variables. \citet{dinh2015nice} consider volume-preserving
transformations which avoid calculating Jacobian determinants.
\citet{salimans2015markov} consider volume-preserving transformations
defined by Markov transition operators.
\citet{rezende2015variational} consider a slightly broader class of
parametric transformations, with Jacobian determinants having at most
$\mathcal{O}(d)$ complexity.

Instead
of specifying a parametric class of mappings, the \gls{VGP} posits a
Bayesian nonparametric prior over all continuous mappings. The
\gls{VGP} can recover a certain class of parametric transformations by
using kernels which induce a prior over that class.
In the context of the \gls{VGP}, the
\gls{GP} is an infinitely wide feedforward network which warps latent
inputs to mean-field parameters. Thus, the \gls{VGP} offers complete
flexibility on the space of mappings---there are no
restrictions such as invertibility or linear complexity---and is fully
Bayesian.
Further, it is a hierarchical variational model, using the \gls{GP} as
a variational prior over mean-field parameters
\citep{ranganath2015hierarchical}. This enables inference over both
discrete and continuous latent variable models.

In addition to its flexibility over parametric methods,
the \gls{VGP} is more computationally efficient. Parametric methods must consider transformations with
Jacobian determinants of at most $\mathcal{O}(d)$ complexity. This
restricts the flexibility of the mapping and therefore the flexibility
of the variational model~\citep{rezende2015variational}. In comparison, the distribution of outputs
using a \gls{GP} prior does not require any Jacobian determinants
(following \myeqp{gp_posterior}); instead it requires auxiliary
inference for inferring variational latent variables (which is fast). Further, unlike discrete Bayesian
nonparametric priors such as an infinite mixture of mean-field
distributions, the \gls{GP} enables black box inference with
lower variance gradients---it applies a location-scale transform for
reparameterization and has analytically tractable KL terms.

Transformations, which convert samples from a tractable distribution
to the posterior, is a classic technique in Bayesian inference. It was
first studied in Monte Carlo methods, where it is core to the
development of methods such as path sampling, annealed importance
sampling, and sequential Monte
Carlo~\citep{gelman1998simulating,neal1998annealed,chopin2002sequential}.
These methods can be recast as specifying a discretized mapping
$f_t$ for times $t_0<\ldots<t_k$, such that for draws $\mbxi$ from the
tractable distribution, $f_{t_0}(\mbxi)$ outputs the same samples and
$f_{t_k}(\mbxi)$ outputs exact samples following the posterior.  By
applying the sequence in various forms,
the transformation bridges the tractable distribution to the
posterior.  Specifying a good transformation---termed ``schedule'' in
the literature---is crucial to the efficiency of these methods. Rather than specify it
explicitly, the
\gls{VGP} adaptively learns this transformation and avoids
discretization.

Limiting the \gls{VGP} in various ways recovers well-known probability models
as variational approximations. Specifically,
we recover the discrete mixture of mean-field
distributions~\citep{bishop1998approximating,jaakola1998improving}.
We also recover a form of factor analysis~\citep{tipping1999probabilistic} in
the variational space. Mathematical details are in
\myappendix{special}.

\section{Experiments}
\label{sec:experiments}

Following standard benchmarks for variational inference in deep
learning, we learn generative models of images. In particular, we
learn the \gls{DLGM}~\citep{rezende2014stochastic}, a
layered hierarchy of Gaussian random variables following neural network
architecures, and the recently proposed
\gls{DRAW}~\citep{gregor2015draw}, a latent attention model that iteratively
constructs complex images using a recurrent architecture and
a sequence of variational auto-encoders~\citep{kingma2014autoencoding}.

For the learning rate we apply a version of
RMSProp~\citep{tieleman2012rmsprop}, in which we scale the value with
a decaying schedule $1/t^{1/2+\epsilon}$ for
$\epsilon>0$. We fix the
size of variational data to be $500$ across all experiments
and set the latent input dimension equal to the number of
latent variables.

\subsection{Binarized MNIST}

\begin{table}[tb]
\centering
\begin{tabular}{lcc}
\toprule
Model & $-\log p(\mbx)$ & $\le$
\\
\midrule
DLGM + {VAE} [\textcolor{MidnightBlue}{1}] &    & 86.76\\
{DLGM} + HVI (8 leapfrog steps) [\textcolor{MidnightBlue}{2}] & 85.51
& 88.30\\
{DLGM} + NF ($k=80$) [\textcolor{MidnightBlue}{3}] &    & 85.10\\
EoNADE-5 2hl (128 orderings) [\textcolor{MidnightBlue}{4}] & 84.68\\
DBN 2hl [\textcolor{MidnightBlue}{5}] & $84.55$\\
DARN 1hl [\textcolor{MidnightBlue}{6}] & $84.13$\\
Convolutional VAE + HVI [\textcolor{MidnightBlue}{2}] & 81.94 & 83.49\\
{DLGM} 2hl + IWAE ($k=50$) [\textcolor{MidnightBlue}{1}] &    & 82.90\\
DRAW [\textcolor{MidnightBlue}{7}] &    & 80.97\\
\midrule
{DLGM} 1hl + \gls{VGP} &    & 84.79\\
{DLGM} 2hl + \gls{VGP} &    & 81.32\\
DRAW + \gls{VGP} &    & \textbf{79.88}\\
\bottomrule
\end{tabular}
\caption{Negative predictive log-likelihood for binarized MNIST.
Previous best results are
[1]~\citep{burda2015importance},
[2]~\citep{salimans2015markov},
[3]~\citep{rezende2015variational},
[4]~\citep{raiko2014iterative},
[5]~\citep{murray2009evaluating},
[6]~\citep{gregor2014deep},
[7]~\citep{gregor2015draw}%
.}
\label{table:mnist}
\end{table}

The binarized MNIST data set~\citep{salakhutdinov2008quantitative}
consists of 28x28 pixel images with binary-valued outcomes.  Training
a \gls{DLGM}, we apply two stochastic layers of 100 random variables
and 50 random variables respectively, and in-between each stochastic
layer is a deterministic layer with 100 units using tanh
nonlinearities. We apply mean-field Gaussian distributions for the
stochastic layers and a Bernoulli likelihood. We train the \gls{VGP}
to learn the \gls{DLGM} for the cases of one stochastic layer and
two stochastic layers.

For \gls{DRAW}~\citep{gregor2015draw}, we augment the
mean-field Gaussian distribution originally used to generate the latent samples
at each time step with the \gls{VGP}, as it places a complex
variational prior over its
parameters. The encoding
recurrent neural network now outputs variational data (used for the
variational model) as well as mean-field
Gaussian parameters (used for the auxiliary model). We use the same architecture hyperparameters as in
\citet{gregor2015draw}.

After training we evaluate test set log likelihood, which are lower
bounds on the true value. See \mytable{mnist} which reports both
approximations and lower bounds of $\log p(\mbx)$ for various methods.
The \gls{VGP} achieves the highest known results on
log-likelihood using \gls{DRAW}, reporting a value of \textbf{-79.88}
compared to the original highest of {-80.97}.
The \gls{VGP} also achieves the highest known results among the class of
non-structure exploiting models using the \gls{DLGM}, with a value of
{-81.32} compared to the previous best of {-82.90} reported by \citet{burda2015importance}.

\subsection{Sketch}
\begin{figure}[t]
\begin{minipage}{\textwidth}
\begin{minipage}[!t]{.45\textwidth}
\centering
\begin{tabular}{lll}
\toprule
Model & Epochs & $\le -\log p(\mbx)$
\\
\midrule
\gls{DRAW} & 100 & 526.8\\
           & 200 & 479.1\\
           & 300 & 464.5\\
\gls{DRAW} + \gls{VGP} & 100 & \textbf{460.1}\\
                       & 200 & \textbf{444.0}\\
                       & 300 & \textbf{423.9}\\
\bottomrule
\end{tabular}
\captionof{table}{Negative predictive log-likelihood for Sketch, learned over hundreds
of epochs over all 18,000 training examples.}
\label{table:sketch}
\end{minipage}
\hfill
\begin{minipage}[!t]{.52\textwidth}
\centering
\includegraphics[width=\columnwidth]{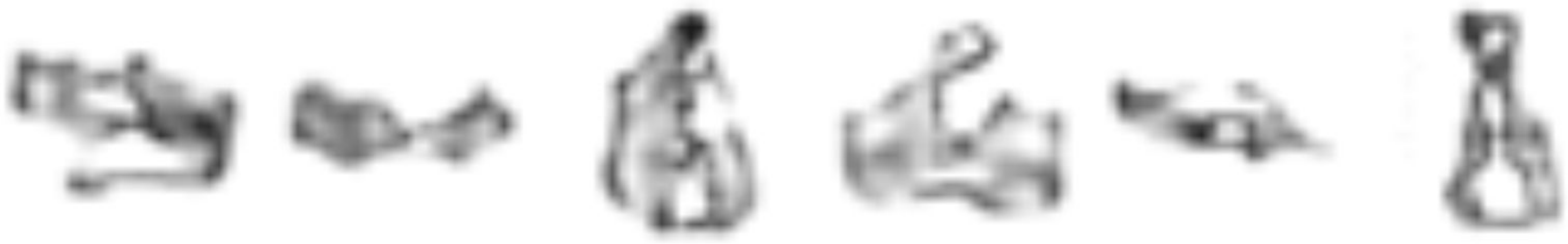}
\includegraphics[width=\columnwidth]{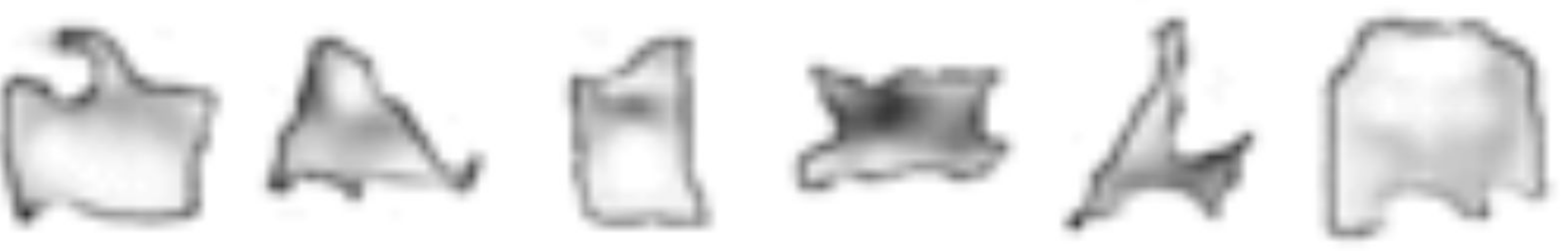}
\captionof{figure}{Generated images from \gls{DRAW} with a \gls{VGP} (top), and
\gls{DRAW} with the original variational auto-encoder (bottom). The
\gls{VGP} learns texture and sharpness, able to sketch more complex shapes.}
\label{fig:sketch}
\end{minipage}
\end{minipage}
\end{figure}
As a demonstration of the \gls{VGP}'s complexity for learning
representations, we also examine
the Sketch data set~\citep{eitz2012hdhso}. It consists of 20,000 human
sketches equally distributed over 250 object categories. We partition
it into 18,000 training examples and 2,000 test examples. We fix
the architecture of \gls{DRAW} to
have a 2x2 read window, 5x5 write attention window, and
64 glimpses---these values were selected using a coarse grid search
and choosing the set which lead to the best training log likelihood.
For inference we use the original auto-encoder version as well as the
augmented version with the \gls{VGP}.

See \mytable{sketch}. \gls{DRAW} with the \gls{VGP}
achieves a significantly better lower bound, performing better than
the original version which has seen state-of-the-art success in many computer
vision tasks. (Until the results presented here, the results from the
original \gls{DRAW}
were the best reported performance for this data set.). Moreover,
the model inferred using the \gls{VGP} is able to generate more
complex images than the original version---it not only performs better
but maintains higher visual fidelity.


%
%
%


\section{Discussion}
\label{sec:discussion}

We present the \acrfull{VGP}, a variational model which adapts its
shape to match complex posterior distributions.
The \gls{VGP}
draws samples from a tractable distribution, and posits a Bayesian
nonparametric prior over transformations from the tractable
distribution to mean-field parameters.
The \gls{VGP} learns the transformations from the space of
all continuous mappings---it is a universal approximator and finds
good posterior approximations via optimization.

In future work the \gls{VGP} will be explored for application in Monte
Carlo methods, where it may be an efficient proposal distribution for
importance sampling and sequential Monte Carlo. An important avenue of
research is also to characterize local optima inherent to the
objective function. Such analysis will improve our
understanding of the limits of the optimization procedure and thus the
limits of variational inference.

\subsubsection*{Acknowledgements}
We thank David Duvenaud, Alp Kucukelbir, Ryan Giordano, and the
anonymous reviewers for their helpful comments.
This work is supported by NSF IIS-0745520, IIS-1247664, IIS-1009542, ONR N00014-11-1-0651, DARPA FA8750-14-2-0009, N66001-15-C-4032, Facebook, Adobe, Amazon, and the Seibel and John Templeton Foundations.

\bibliographystyle{iclr2016_conference}
\bibliography{iclr2016}

\appendix

\section{Special cases of the variational Gaussian process}
\label{appendix:special}


We now analyze two special cases of the \gls{VGP}: by limiting its
generative process in various ways, we recover well-known models.
This
provides intuition behind the \gls{VGP}'s complexity.  In
\mysec{related} we show many recently proposed models can
also be viewed as special cases of the \gls{VGP}.

\begin{example}
A mixture of mean-field distributions is a \gls{VGP} without a
kernel.
\end{example}

A discrete mixture of mean-field
distributions~\citep{bishop1998approximating,jaakola1998improving,lawrence2000variational}
is a classically studied variational model
with dependencies between latent variables.
Instead of a mapping which interpolates between
inputs of the variational data,
suppose the \gls{VGP}
simply performs nearest-neighbors for a latent input
$\mbxi$---selecting
the output $t_n$ tied to the nearest variational input $s_n$.
This induces a multinomial distribution of outputs, which samples one
of the variational outputs' mean-field parameters.%
\footnote{%
Formally, given variational input-output pairs $\{(\mbs_n,\mbt_n)\}$,
the nearest-neighbor function is defined
as $f(\mbxi)=\mbt_j$, such that
$\|\mbxi- \mbs_j\|<\|\mbxi- \mbs_k\|$ for all $k$.
Then the output's distribution is multinomial with
probabilities $P(f(\mbxi)=\mbt_j)$, proportional to areas of the
partitioned nearest-neighbor space.
}
Thus, with a \gls{GP} prior that interpolates between
inputs, the \gls{VGP} can be seen as a kernel density smoothing of the
nearest-neighbor function.

\begin{example}
Variational factor analysis is
a \gls{VGP} with linear kernel and no variational data.
\end{example}

Consider factor analysis~\citep{tipping1999probabilistic} in the
variational space:
\footnote{\label{footnote:degenerate_mf}
For simplicity, we avoid discussion of the \gls{VGP}'s underlying
mean-field distribution, i.e., we specify each mean-field factor
to be a degenerate point mass at its parameter value.
}
\begin{equation*}
\mbxi \sim \mathcal{N}(\mbzero, \mbI),
\qquad
\mbz_i \sim \mathcal{N}(\mbf{w}^\top\mbxi, \mbI).
\end{equation*}
Marginalizing over the latent inputs induces
linear dependence in $\mbz$,
$q(\mbz;\mbf{w})=\mathcal{N}(\mbz; \mbzero,\mbf{w}\mbf{w}^\top)$.
Consider the dual interpretation
\begin{equation*}
\mbxi \sim \mathcal{N}(\mbzero, \mbI),
\qquad
f_i \sim \mathcal{GP}(0,k(\cdot,\cdot)),
k(\mbs,\mbs')=\mbs^\top\mbs',
\qquad
\mbz_i=f_i(\mbxi),
\end{equation*}
with
$q(\mbz\g\mbxi) = \mathcal{N}(\mbz; \mbzero,\mbxi\mbxi^\top)$.
The maximum likelihood estimate of $\mbf{w}$ in factor analysis is the
maximum a posteriori estimate of $\mbxi$ in the \gls{GP} formulation.
More generally, use of a non-linear kernel induces non-linear
dependence in $\mbz$.
Learning the set of kernel hyperparameters $\mbtheta$ thus learns the set
capturing the most variation in its
latent embedding of $\mbz$~\citep{lawrence2005probabilistic}.


\section{Proof of \mytheorem{limit}}
\label{appendix:zero}

\begingroup
\def\thetheorem{\ref{theorem:limit}}
\begin{theorem}
\limittheorem
\end{theorem}
\addtocounter{theorem}{-1}
\endgroup
\begin{proof}
Let the mean-field distribution be given by degenerate delta distributions
\begin{align*}
q(\mbz_i \g f_i) = \delta_{f_i}(\mbz_i).
\end{align*}
Let the size of the latent input be equivalent to the number of
latent variables $c=d$ and fix $\sigma^2_{\textsc{ard}}=1$ and $\mbomega_j=1$. Furthermore for simplicity, we assume that
$\mbxi$ is drawn uniformly on the $d$-dimensional hypercube. Then
as explained in \mysec{vgp:universal},
if we let $P^{-1}$ denote the inverse posterior cumulative distribution
function, the optimal $f$ denoted $f^*$ such that
\begin{equation*}
\operatorname{KL}(q(\mbz; \mbtheta) \gg p(\mbz\g\mbx))
= 0
\end{equation*}
is
\begin{align*}
f^*(\xi) = P^{-1}(\mbxi_1, ..., \mbxi_d).
\end{align*}
Define ${\cal O}_k$ to be the set of points $j/2^k$ for $j = 0$ to $2^k$, and define ${\cal S}_k$ to be the $d$-dimensional product of ${\cal O}_k$. Let
$\cD_k$ be the set containing the pairs $(s_i, f^*(s_i))$, for each element $s_i$ in ${\cal S}_k$. Denote $f^k$ as the
\acrshort{GP} mapping conditioned on the dataset $\cD_k$, this
random mapping satisfies $f^k(s_i) = f^*(s_i)$ for all $s_i \in {\cal S}_k$ by the noise free prediction
property of Gaussian processes~\citep{rasmussen2006gaussian}.
Then by continuity, as $k \to \infty$, $f^k$  converges to $f^*$.
\end{proof}
A broad condition under which the quantile function of a distribution
is continuous is if that distribution has positive density with
respect to the Lebesgue measure.

The rate of convergence for finite sizes of the variational data
can be studied via posterior contraction rates for \glspl{GP} under random
covariates~\citep{van2011information}. Only an additional assumption using
stronger continuity conditions for the posterior quantile and the use
of Matern covariance functions is required for the theory to be applicable in the variational setting.


\section{Variational objective}
\label{appendix:variational}

We derive the tractable lower bound to the model evidence $\log
p(\mbx)$ presented in \myeqp{variational_objective}.
To do this, we first penalize the
\gls{ELBO} with an expected KL term,
\begin{align*}
\log p(\mbx)
&\ge \mathcal{L}
=
\mathbb{E}_{\qvgp}[\log p(\mbx\g\mbz)]
- \operatorname{KL}(\qvgp(\mbz)\|p(\mbz))
\\
&\ge
\mathbb{E}_{\qvgp}[\log p(\mbx\g\mbz)]
- \operatorname{KL}(\qvgp(\mbz)\|p(\mbz))
- \mathbb{E}_{\qvgp}\Big[
\operatorname{KL}(q(\mbxi,f\g\mbz)\|r(\mbxi,f\g\mbz))
\Big].
\end{align*}
We can combine all terms into the expectations as follows:
\begin{align*}
\widetilde{\cL}
&=
\mathbb{E}_{q(\mbz,\mbxi,f)}\Big[
\log p(\mbx\g \mbz) - \log q(\mbz)
+\log p(\mbz)
-
\log q(\mbxi,f\g\mbz) + \log r(\mbxi,f\g\mbz)
\Big]
\\
&=
\mathbb{E}_{q(\mbz,\mbxi,f)}\Big[
\log p(\mbx\g \mbz) - \log q(\mbz\g f(\mbxi))
+\log p(\mbz)
-
\log q(\mbxi,f) + \log r(\mbxi,f\g\mbz)
\Big]
,
\end{align*}
where we apply the product rule
$q(\mbz)q(\mbxi,f\g\mbz)=q(\mbz\g f(\mbxi))q(\mbxi,f)$.
Recombining terms as KL divergences, and written with parameters
$(\mbtheta,\mbphi)$, this recovers the auto-encoded variational
objective in \mysec{bbi}:
\begin{align*}
\widetilde{\cL}(\mbtheta,\mbphi)
&
=
\mathbb{E}_{\qvgp}[\log p(\mbx\mid\mbz)]
-
\mathbb{E}_{\qvgp}\Big[
\operatorname{KL}(q(\mbz\g f(\mbxi))\|p(\mbz))
\Big]
\\
&
\quad
-
\mathbb{E}_{\qvgp}\Big[
\operatorname{KL}(q(f\g\mbxi;\mbtheta)\|r(f\g\mbxi,\mbz;\mbphi))
+
\log q(\mbxi) - \log r(\mbxi\g\mbz)
\Big].
\end{align*}
%
%
The KL divergence between the mean-field $q(\mbz\g f(\mbxi))$ and the
model prior $p(\mbz)$ is analytically tractable for certain popular
models.
For example, in the deep latent Gaussian
model~\citep{rezende2014stochastic} and \gls{DRAW}~\citep{gregor2015draw}, both the mean-field
distribution and model prior are Gaussian, leading to an analytic KL
term:
for Gaussian random variables of dimension $d$,
\begin{align*}
\operatorname{KL}&(\mathcal{N}(\mbx; \m_1,\mbSigma_1)\|\mathcal{N}(\mbx; \m_2,\mbSigma_2))
=
\\
&\frac{1}{2}
\left(
(\m_1-\m_2)^\top\mbSigma_1^{-1}(\m_1-\m_2) +
\operatorname{tr}(\mbSigma_1^{-1}\mbSigma_2 + \log{\mbSigma_1} -
\log{\mbSigma_2}) - d\right).
\end{align*}
In general, when the KL is intractable,
we combine the KL term with the reconstruction term, and maximize the
variational objective
\begin{align}
\begin{split}
\widetilde{\cL}(\mbtheta,\mbphi)
&
=
\mathbb{E}_{\qvgp}[\log p(\mbx,\mbz) - \log q(\mbz\g f(\mbxi))]
\\
&
\quad
-
\mathbb{E}_{\qvgp}\Big[
\operatorname{KL}(q(f\g\mbxi;\mbtheta)\|r(f\g\mbxi,\mbz;\mbphi))
+
\log q(\mbxi) - \log r(\mbxi\g\mbz)
\Big].
\end{split}
\label{eq:general_vgp_bound}
\end{align}
We expect that this experiences slightly higher variance in the
stochastic gradients during optimization.

We now consider the second term.
Recall that we specify the auxiliary model to be a fully factorized
Gaussian,
$r(\mbxi,f\g\mbz) = \mathcal{N}((\mbxi,f(\mbxi))^\top\g\mbz; \m,\S)$,
where $\m\in\mathbb{R}^{c+d}$, $\S\in\mathbb{R}^{c+d}$. Further, the
variational priors $q(\mbxi)$ and $q(f\g\mbxi)$ are both defined to be
Gaussian. Therefore it is also a KL
divergence between Gaussian distributed
random variables. Similarly, $\log q(\mbxi) - \log r(\mbxi\g\mbz)$ is
simply a difference of Gaussian log densities.
%
The second expression is simple to compute and backpropagate gradients.

\section{Gradients of the variational objective}
\label{appendix:gradients}

We derive gradients for the variational objective
(\myeqp{reparam_objective}). This follows trivially by backpropagation:
\begin{align*}
\nabla_{\mbtheta}
\widetilde{\cL}(\mbtheta, \mbphi)
&=
\mathbb{E}_{\mathcal{N}(\mbxi)}[
\mathbb{E}_{w(\mbepsilon)}[
\nabla_{\mbtheta} \f(\mbxi)\nabla_{\f}\mbz(\mbepsilon)\nabla_{\mbz}
\log p(\mbx\g\mbz)
]]
\\
&\quad
-
\mathbb{E}_{\mathcal{N}(\mbxi)}\Big[\mathbb{E}_{w(\mbepsilon)}\Big[
\nabla_{\mbtheta}
\operatorname{KL}(q(\mbz\g \f(\mbxi;\mbtheta))\|p(\mbz))
\Big]\Big]
\\
&\quad
-
\mathbb{E}_{\mathcal{N}(\mbxi)}\Big[\mathbb{E}_{w(\mbepsilon)}\Big[
\nabla_{\mbtheta}
\operatorname{KL}(q(f\g\mbxi;\mbtheta)\|r(f\g\mbxi,\mbz;\mbphi))
\Big]\Big]
,
\\
\nabla_{\mbphi}
\widetilde{\cL}(\mbtheta, \mbphi)
&=
-
\mathbb{E}_{\mathcal{N}(\mbxi)}[\mathbb{E}_{w(\mbepsilon)}[
\nabla_{\mbphi}
\operatorname{KL}(q(f\g\mbxi;\mbtheta)\|r(f\g\mbxi,\mbz;\mbphi))
-
\nabla_{\mbphi}
\log r(\mbxi\g\mbz; \mbphi)
]
],
\end{align*}
where we assume the KL terms are analytically written from
\myappendix{variational} and gradients are propagated similarly through their
computational graph.
In practice, we need only be
careful about the expectations, and the gradients of the functions
written above are taken care of with automatic differentiation tools.

We also derive gradients for the general variational bound of
\myeqp{general_vgp_bound}---it assumes that the first KL term,
measuring the divergence between $q$ and the prior for $p$, is not
necessarily tractable.
Following the reparameterizations described in \mysec{bbi:gradient},
this variational objective can be rewritten as
\begin{align*}
\widetilde{\cL}(\mbtheta, \mbphi)
&=
\mathbb{E}_{\mathcal{N}(\mbxi)}\Big[\mathbb{E}_{w(\mbepsilon)}\Big[
\log p(\mbx,\mbz(\mbepsilon;\f)) - \log
q(\mbz(\mbepsilon;\f)\g\f)
\Big]\Big]
\\
&\quad
-
\mathbb{E}_{\mathcal{N}(\mbxi)}\Big[\mathbb{E}_{w(\mbepsilon)}\Big[
\operatorname{KL}(q(f\g\mbxi;\mbtheta)\|r(f\g\mbxi,\mbz(\mbepsilon;\f);\mbphi))
+
\log q(\mbxi) - \log r(\mbxi\g\mbz(\mbepsilon;\f))
\Big]\Big]
.
\end{align*}
We calculate gradients by backpropagating over the nested
reparameterizations:
\begin{align*}
\nabla_{\mbtheta}
\widetilde{\cL}(\mbtheta, \mbphi)
&=
\mathbb{E}_{\mathcal{N}(\mbxi)}[
\mathbb{E}_{w(\mbepsilon)}[
\nabla_{\mbtheta} \f(\mbxi)
\nabla_{\f} \mbz(\mbepsilon)[\nabla_{\mbz}\log p(\mbx,\mbz) - \nabla_{\mbz}\log q(\mbz\g\f)]]]
\\
&\quad
-
\mathbb{E}_{\mathcal{N}(\mbxi)}\Big[\mathbb{E}_{w(\mbepsilon)}\Big[
\nabla_{\mbtheta}
\operatorname{KL}(q(f\g\mbxi;\mbtheta)\|r(f\g\mbxi,\mbz;\mbphi))
\Big]\Big]
\\
\nabla_{\mbphi}
\widetilde{\cL}(\mbtheta, \mbphi)
&=
-
\mathbb{E}_{\mathcal{N}(\mbxi)}[\mathbb{E}_{w(\mbepsilon)}[
\nabla_{\mbphi}
\operatorname{KL}(q(f\g\mbxi;\mbtheta)\|r(f\g\mbxi,\mbz;\mbphi))
-
\nabla_{\mbphi}
\log r(\mbxi\g\mbz; \mbphi)
]
]
.
\end{align*}

\section{Scaling the size of variational data}
\label{appendix:size}

If massive sizes of variational data are required, e.g., when its
cubic complexity due to inversion of a $m\times m$ matrix becomes the
bottleneck during computation, we can scale it further.
Consider fixing the variational inputs to lie on a grid.
For stationary kernels, this allows us to exploit Toeplitz structure
for fast $m\times m$ matrix inversion. In particular, one can embed
the Toeplitz matrix into a circulant matrix and apply conjugate
gradient combined with fast Fourier transforms in order to compute
inverse-matrix vector products in $\mathcal{O}(m\log m)$ computation
and $\mathcal{O}(m)$ storage~\citep{cunningham2008fast}.
For product kernels, we can further exploit Kronecker structure to
allow fast $m\times m$ matrix inversion in $\mathcal{O}(Pm^{1+1/P})$
operations and $\mathcal{O}(Pm^{2/P})$ storage, where $P>1$ is the
number of kernel products~\citep{osborne2010bayesian}. The \gls{ARD}
kernel specifically leads to $\mathcal{O}(cm^{1+1/c})$ complexity,
which is linear in $m$.

\end{document}